\documentclass[9pt]{article}
\usepackage{CJKutf8}
\usepackage{spconf,amssymb,amsmath,graphicx}
\usepackage{bm}
\usepackage{multirow}
\usepackage{booktabs}
\usepackage{hyperref}
\usepackage[numbers]{natbib}
\usepackage{times}
\usepackage{textcomp}
\usepackage{hyperref}
\hypersetup{colorlinks=true}
\usepackage[final]{changes}

\usepackage{bibspacing}
\usepackage{color}
\definecolor{darkspringgreen}{rgb}{0.09, 0.45, 0.27}

\newcommand{\vect}[1]{\mathbf{#1}}

\title{Massively Multilingual Shallow Fusion with Large Language Models}
\name{\begin{tabular}{c} Ke~Hu,~Tara~N.~Sainath,~Bo~Li,~Nan~Du,~Yanping~Huang,~Andrew~M.~Dai,~Yu~Zhang,\\ Rodrigo~Cabrera,~Zhifeng~Chen,~Trevor~Strohman \end{tabular} \thanks{We thank Maxim Krikun and David So for useful discussions during this work.}}
\address{Google LLC, USA \\
\fontsize{9}{9}\selectfont\ttfamily\upshape
\{huk,tsainath,boboli\}@google.com}
\begin{document}
\ninept
\maketitle
\begin{abstract}
While large language models (LLM) have made impressive progress in natural language processing, it remains unclear how to utilize them in improving automatic speech recognition (ASR). In this work, we propose to train a single multilingual language model (LM) for shallow fusion in multiple languages. We push the limits of the multilingual LM to cover up to 84 languages by scaling up using a mixture-of-experts LLM, i.e., generalist language model (GLaM). When the number of experts increases, GLaM dynamically selects only two at each decoding step to keep the inference computation roughly constant. We then apply GLaM to a multilingual shallow fusion task based on a state-of-the-art end-to-end model. Compared to a dense LM of similar computation during inference, GLaM reduces the WER of an English long-tail test set by 4.4\% relative. In a multilingual shallow fusion task, GLaM improves 41 out of 50 languages with an average relative WER reduction of 3.85\%, and a maximum reduction of 10\%. Compared to the baseline model, GLaM achieves an average WER reduction of 5.53\% over 43 languages.
\end{abstract}
%
%
\section{Introduction \label{sec:intro}}

End-to-end (E2E) models have increased in popularity in recent years, with many research groups showing promising results~\cite{he19streaming, KimHoriWatanabe17, JinyuLi2019, Zeyer2020}. Despite the success of these models, one of the main issues with these models continues to be performance on long-tail named entities. E2E models are most often trained on audio-text pairs, which is a fraction of data compared to the large amount of text-only data available. This fact, coupled with  the limitation that E2E models often run with a small beam size (i.e., 8 hypotheses), contribute to poor performance on the long-tail. 

There has been many efforts on improving long-tail quality, including external language models trained on text-only data~\cite{Chorowski17, Anjuli18}, using text-to-speech systems to convert the text into paired audio-text pairs~\cite{chen2021injecting}, cycle-consistency losses~\cite{hori2019cycle, tjandra2017listening}, or injecting text directly into the E2E model~\cite{bapna2021slam, tang2022unified, thomas2022towards, Zhehuai2022, Sainath2023} -- thus improving ASR performance without increasing model parameters or decoding complexity. So far, combining an external neural language model with the E2E model continues to be one of the most effective methods to incorporate text-only data~\cite{Sainath2023}. However, most of the neural language models are either relative small, on the order of 100M parameters or less~\cite{sainath2021cascadedlm, meng2021minimum, huang2022sentence,inaguma2019transfer}\added{, or focusing on monolingual scenarios~\cite{irie2019language}}.

Recently, in the natural language processing (NLP) community there has been huge success with large language models (LLM), including generative pre-trained transformers 3 (GPT-3)~\cite{brown2020language}, GPT-J~\cite{wang2021gpt}, Meena~\cite{adiwardana2020towards}, LaMDA~\cite{thoppilan2022lamda}, Megatron-Turing natural language generation (MT-NLG) model~\cite{smith2022using}, and GLaM~\cite{du2022glam}, which are on the order of billions to trillions of parameters. In particular, GLaM is a mixture-of-experts based model, and a benefit of GLaM models is that while they are trained with a massive number of parameters, they are sparse so that during inference the computation remains on-par with the smaller, production-size LMs (e.g., $\sim$100M parameters).  However, to date these LLMs have not been explored in the speech recognition domain. In addition, these language models are typically trained with many external sources of data such as webpages, books, Wikipedia pages, forums and news pages~\cite{du2022glam}. These tasks are often not matched in domain to some of the ASR use cases, most notably Voice Search.

In this work, we explore how large language models can be used in an ASR task. Specifically, we focus on the GLaM LLM~\cite{du2022glam}. Our proposed novelties are two folds: First, to our knowledge this is the first application of LLMs into speech recognition. Second, unlike previous LLM work, our LM is trained on domain-specific Voice Search data across many languages.

We investigate the use of the GLaM LLM on a 84-language Voice Search task by using GLaM for shallow fusion with a massively multilingual E2E model~\cite{zhang2021bigssl}. We compare this to fusing with a dense LM of 140M parameters trained on the same data. \added{We show that a 64-expert GLaM reduces the WER for an English long-tail test set by 4.4\% relative. In multilingual shallow fusion, GLaM reduces the WERs for 41 of 50 languages with an average relative WER reduction of 3.85\%, with the reduction up to 10\% relative. Although one may need to store the 1.9G-parameter GLaM on disk, the inference computation of GLaM is comparable to a 140M Conformer LM due to its sparsity. To the best of our knowledge, this is the first time that an LLM is utilized in such a massive scale in terms of languages and model size to improve ASR performance. }
 
\section{Large Language Model Shallow Fusion}
\label{sec:multi_sf}

\subsection{Multilingual Shallow Fusion}

Shallow fusion is a way to incorporate external language models to ASR models during decoding \cite{kannan2018analysis, chorowski2016towards}. Given an input speech sequence $\vect{x}$ and the posterior probability from an E2E model $p(\vect{y}|\vect{x})$, shallow fusion combines the E2E model score with the language model score $p_{LM}(\vect{y})$ at each beam search step:
\begin{equation}
\vect{y}^* = \underset{\vect{y}}{\arg\max} [\log p(\vect{y}|\vect{x}) + \lambda \log p_{LM}(\vect{y})]
\label{eq:sf}
\end{equation}
\noindent
Shallow fusion capitalizes on the LM at each decoding step and has the potential to bring new tokens to the beam. Different from previous work where monolingual LMs are often used \cite{sainath2021efficient, meng2021minimum, irie2019language}, in this work, we train a single language model by combining up to 84 languages for shallow fusion. We do not explicitly use any ground truth language information in training or inference. We explore a 64-expert GLaM model \cite{du2022glam} for shallow fusion and compare to dense LMs such as Conformer \cite{gulati2020conformer}.

\subsection{GLaM}
\added{GLaM \cite{du2022glam} is a type of LLM which utilizes mixture-of-experts (MoE) layers to scale up to trillions of parameters. GLaM achieves better overall zero, one and few-shot performance across 29 NLP tasks than GPT-3 but is more training and inference efficient due to its sparsity. Although LLMs such as GLaM have achieved impressive progress in NLP, their power has not been utilized in ASR. In this work, we incorporate the GLaM LLM in multilingual ASR by shallow fusion.}

Compared to a dense LM such as Conformer \cite{sainath2021efficient, gulati2020conformer}, GLaM is a sparse LM where only parts of the model are activated during forward propagation. \added{When using a large-size GLaM for inference, although we need to store the whole GLaM model on disk, the computation remains efficient due to the sparsity.} For example, a 1.9B parameter GLaM model in \cite{du2022glam} has 64 experts but only activates 2 experts during inference time. Therefore, the computation is similar to a 145M-parameter dense model. GLaM achieves this by using an MoE layer \cite{shazeer2017outrageously} which utilizes a gating network to select a couple of experts from a number of them for computation. In \cite{du2022glam}, the MoE layer replaces the feed forward component of every other transformer layer. Both the MoE layer and the gating network are jointly trained. The selection is dynamic and on the token level. Following \cite{du2022glam}, we dynamically activate two experts in our GLaM LM during shallow fusion.

In our work, we train GLaM to cover 84 languages. The total size of our GLaM model is around 1.9B, and only two experts are activated for inference computation, which is comparable to a 140M Conformer LM \cite{sainath2021efficient} (see Sect. \ref{sec:compare} for details). We combine transcripts from both speech-text data and text-only data from all languages in training to avoid domain mismatch. When training GLaM, we follow \cite{du2022glam} and partition the weights and computation of the GLaM model using the GSPMD algorithm described in \cite{xu2021gspmd}. We use cross-entropy loss and Adafactor \cite{shazeer2018adafactor} optimizer for training.

\section{Experimental Details}
\label{sec:exp}

We perform our experiments using large-scale speech-text and text-only data from 84 languages and a sate-of-the-art E2E baseline model \cite{bigbench2022}.

\subsection{Data}

\subsubsection{Speech-Text Data}

Our multilingual training data come from a total of 84 languages. The list of languages and their numbers of training utterances are illustrated in Fig. \ref{fig:84_lang}(a). The number of utterances for each language ranges from 2K to 66M, with a median number of 8.5M. The speech-text training data are randomly sampled from Google Voice Search traffic and are de-identified and human transcribed, and abide by Google AI Principle \cite{googleaiprinciples}. Due to test data availability at the time of this research, we only have test sets for 50 languages. 
The number of test utterances range from 1.5K to 19.7K utterances per language, and they are sampled from Voice Search traffic and anonymized and human transcribed for evaluation purposes.

\subsubsection{Text-Only Data}
\begin{figure*}[t]
  \centering
   \includegraphics[scale=0.19]{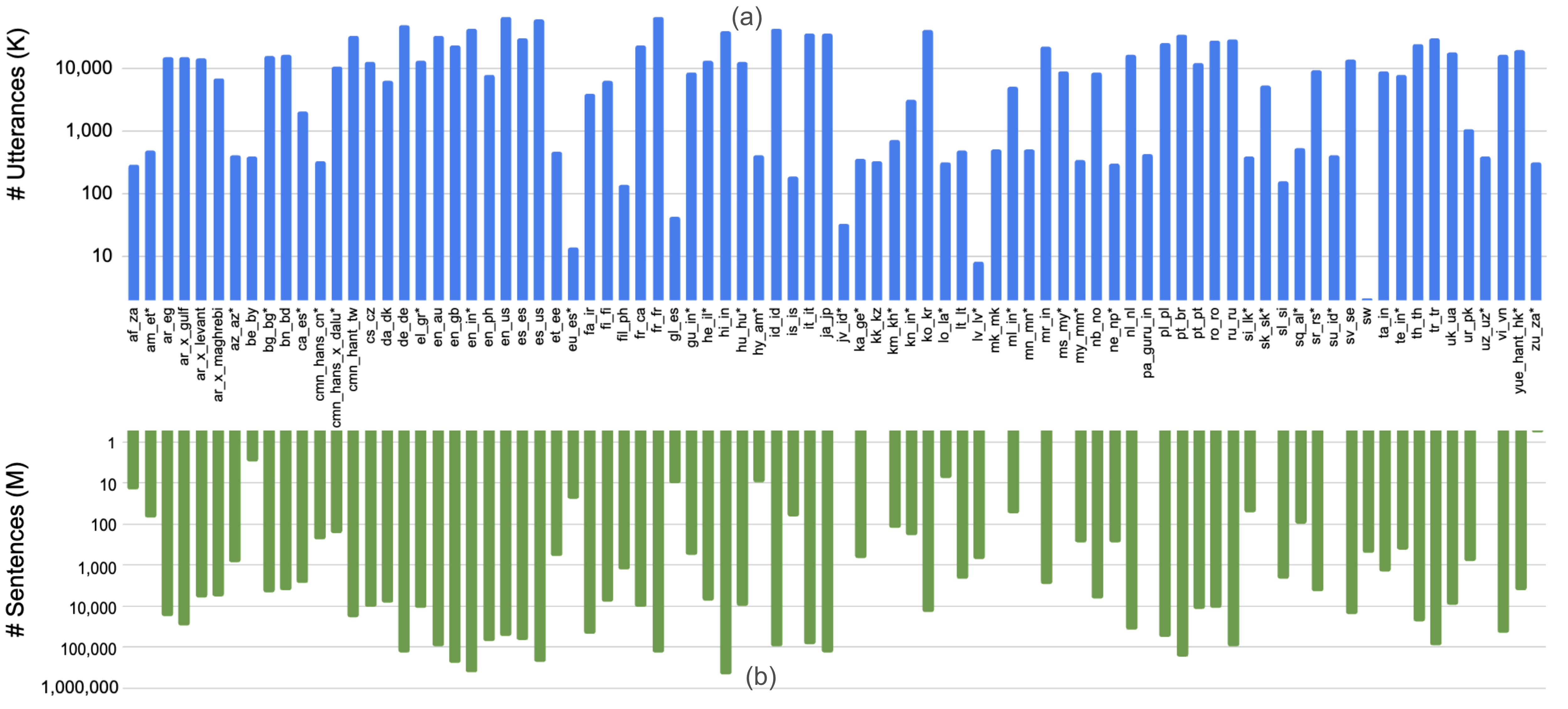}
   \caption{Training data distribution for 84 languages from different locales. There are two types of data, i.e., (a) speech-text data, and (b) text-only data. Note that the utterance and sentence counts are shown on a logarithmic scale, and may be missing for some languages. Languages marked with asterisks (*) are the ones without test sets. The ``x" in language names means multiple regions in the area. For example, cmn\_hans\_x\_dalu means Mandarin Chinese (Simplified) in multiple regions in China mainland.}
   \label{fig:84_lang}
\end{figure*}

Besides speech-text data, we add text-only data for language model training. Our text data covers 75 of the 84 languages, and the list of languages and the per-language numbers of sentences are illustrated in Fig. \ref{fig:84_lang}(b). The sentences for each language ranges from 581K to 451B,  Around 50\% of languages have more than 1B sentences. The sentences are sampled from anonymized search traffic across multiple domains such as Web, Maps, News, Play, and YouTube. 

\subsection{Model Architecture}

\subsubsection{E2E Model}
In the large-scale shallow fusion experiments, we use a 2B-parameter E2E attention model \cite{zhang2021bigssl} as the baseline model. 128D log-mel filterbank features are extracted from each frame and then stacked together from 3 continuous frames to form a 512D input vector. These input vectors are further downsampled to have a 30-ms frame rate. Similar to \cite{zhang2021bigssl}, our baseline model is first pretrained by using YouTube-based large-scale data and then fine tuned using the speech-text data from the 84 languages in Fig. \ref{fig:84_lang}(a). The E2E model is an attention model consisting of a 32-layer 1536-D Conformer encoder and a 6-layer 768-D long-short term memory (LSTM) decoder, with a total of of 2B parameters. The Conformer layer has a kernel size of 5, and both a left context and right context of 3.87 sec. A softmax is used to predict 16,384 wordpieces. We generate the wordpieces using mixed transcripts pooled from all languages. We use the GShard \cite{lepikhin2020gshard} framework with the GSPMD backend \cite{xu2021gspmd} to train the 2B model on Google Cloud V4 TPUs. The model is trained with a transformer learning rate schedule with a peak learning rate of 2.5e-3 and a linear warm-up phase of 5K steps.

\subsubsection{GLaM Architecture and Training}
\label{sec:glam_architecture}

Our GLaM model has 12 transformer layers with a model dimension of 768. It consists of 64 experts, and we use 12-headed self-attention with a hidden dimension of 64. For input data, we pack training sentences into a batch example using a factor of 4 and maximum sequence length of 1024. The GLaM LM is trained by using the text portion of the speech-text data as well as the text-only data of 84 languages in Fig. \ref{fig:84_lang}. The total size of the GLaM model is 1.9B. However, in inference we only select the top 2 experts for forward propagation, which in total amounts to less than 8\% of the network, i.e., around 145M parameters \cite{du2022glam}. We follow \cite{du2022glam} to train the GLaM model. We partition the weights and computation of the GLaM model using the GSPMD algorithm as described in \cite{xu2021gspmd}. We use Adafactor \cite{shazeer2018adafactor} optimizer with first-moment decay $\beta_1=0$, second-moment decay $\beta_2=0.99$, a decay schedule of $1-t^{-0.8}$, update clipping threshold of 1.0, and factored second-moment estimation. We train the GLaM until the loss is relatively stable. The GLaM model is trained on 128 Cloud TPU-V4 chips.

\section{Results and Discussion}

\begin{figure*}[t]
  \centering
   \includegraphics[scale=0.19]{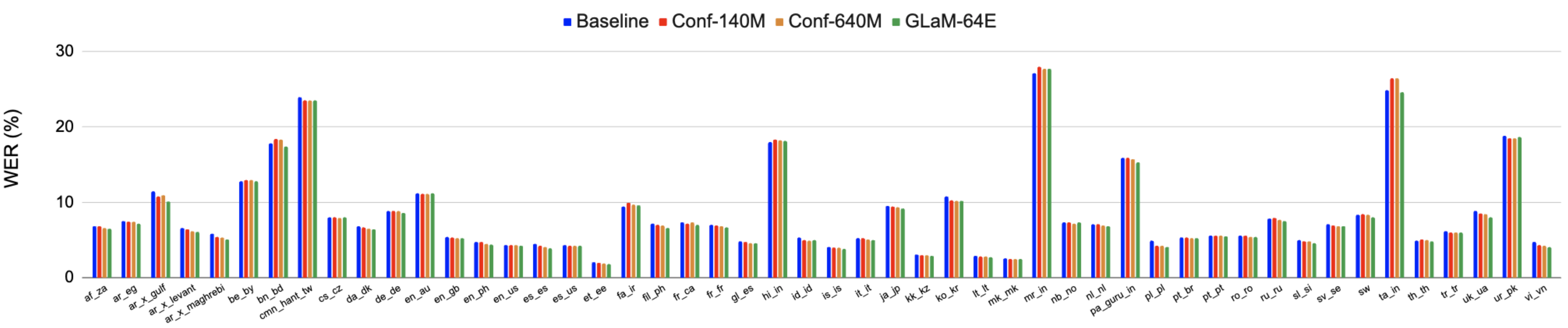}
   \caption{WER comparison between the baseline teacher model and shallow fusion using different language models.}
   \label{fig:wer}
\end{figure*}

\begin{figure*}[t]
  \centering
   \includegraphics[scale=0.19]{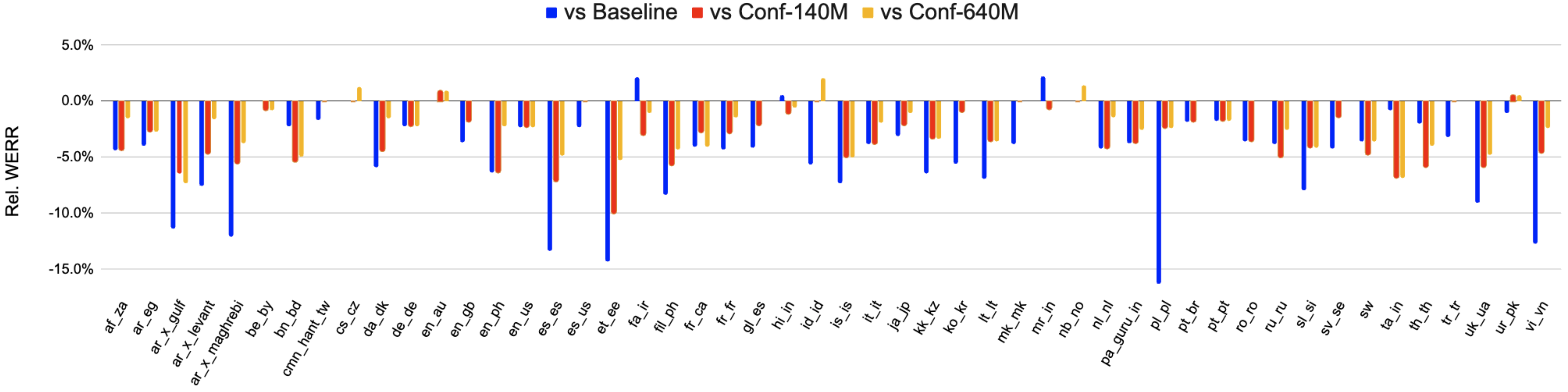}
   \caption{Relative WER reduction (WERR) by using the GLaM language model for shallow fusion compared to the baseline teacher model and shallow fusion based on Conformer language models.}
   \label{fig:werr}
\end{figure*}

\subsection{Results on English}

We first evaluate shallow fusion using the English Voice Search (VS) test set and a long-tail test set. The long-tail test set is generated by synthesized utterances using text sentences from the Maps domain containing proper nouns rare in speech-text training data. This Rare Proper Noun Maps (RPNM) set consists of 10K utterances in total. We use the 64-expert GLaM, as well as Conformer LMs of two different sizes, i.e., 140M and 640M, respectively, for comparison. To understand the effect of scaling up GLaM, we also show results of a 2-expert GLaM, which is a special dense version of the MoE model. All LMs are trained using the same data.

The results are shown in Table \ref{tab:teacher_en}. Compared to the baseline, the 140M Conformer LM (Conf-140M) improves the RPNM WER by 3.4\% relative and leaves en-US VS unchanged. This is consistent with the result that LMs often improves long-tail performance significantly \cite{sainath2021efficient, hu2022improving}. When the Conformer LM size increases to 640M (Conf-640M), the RPNM WER reduction improves to 6.0\% relative, showing the effectiveness of increasing model capacity in LM modeling. The 2-expert GLaM (GLaM-2E) model performs similarly to a 140M Conformer LM because a 2-expert GLaM reduces to a dense model. When we increase the number of experts to 64 (GLaM-64E), we see that GLaM improves en-US VS WER slightly (0.1\% absolute) and RPNM by 4.6\% relative, compared to Conf-140M. The improvement is due to the use of multiple experts. We will use GLaM-64E for the following large-scale experiment.

\begin{table}[t]
\centering
\begin{tabular}{lcc}
    \toprule
    \multirow{2}{*}{Shallow Fusion} & \multicolumn{2}{c}{WER (\%)} \\ \cline{2-3}
    & \multirow{1.2}{*}{\shortstack{en-US VS}} & \multirow{1.2}{*}{RPNM} \\ \midrule \midrule
    2B Baseline & 4.3 & 11.7 \\
    Conf-140M & 4.3 & 11.3 \\
    Conf-640M & 4.3 & 11.0 \\
    GLaM-2E & 4.3 & 11.2 \\
    GLaM-64E & \textbf{4.2} & \textbf{10.8} \\ \hline
\end{tabular}
\caption{WERs (\%) for en-US VS and RPNM between the baseline teacher and shallow fusion using different language models.}
\label{tab:teacher_en}
\end{table}

\begin{table*}[h]
\footnotesize
\centering
\begin{tabular}{ |c|c|c|c| }
    \hline
    & Language & GLaM-64E & Conformer-140M \\ \hline
    \multirow{7}{*}{Wins} & \multirow{2}{*}{es\_es} & {\color{darkspringgreen} edad de serena y venus williams} & serena {\color{red} ibrahimovic} \\ \cline{3-4}
     & & elsa {\color{darkspringgreen} pataky} & elsa {\color{red} patati} \\ \cline{2-4}
     & \multirow{2}{*}{en\_ph} & search {\color{darkspringgreen} ni yao de ai} english version & search {\color{red} ne-yo the eye} english version \\ \cline{3-4}
     & & who's the member of {\color{darkspringgreen} blackpink} & who's the member of {\color{red} black pink} \\ \cline{2-4}
     & \multirow{1}{*}{nl\_nl} & {\color{darkspringgreen} peter van de veire} & {\color{red} pieter vandevere} \\ \cline{2-4}
     & \multirow{1}{*}{cmn\_hant\_tw} & {\color{darkspringgreen} \begin{CJK*}{UTF8}{bsmi} 馮 提 莫 \end{CJK*}} & {\color{red} pokemon} \\ \hline
    \multirow{3}{*}{Losses} & \multirow{1}{*}{en\_au} & {\color{red} what do you} want to play & {\color{darkspringgreen} nothing to} want to play \\ \cline{2-4}
    & \multirow{1}{*}{nl\_nl} & {\color{red} onze-lieve-vrouwstraat} 7 oostakker & {\color{darkspringgreen} onze lieve vrouwstraat} 7 oostakker \\ \cline{2-4}
    & \multirow{1.1}{*}{cmn\_hant\_tw} & \begin{CJK*}{UTF8}{bsmi} 華 冠 \end{CJK*} {\color{red} sapporo} & \begin{CJK*}{UTF8}{bsmi} 華 冠 \end{CJK*}  {\color{darkspringgreen} \begin{CJK*}{UTF8}{bsmi} 三 杯 肉 \end{CJK*}} \\ \hline
\end{tabular}
\caption{Decoding examples of GLaM and Conformer shallow fusion. Wins are in green and losses in red.}
\label{tab:decoding_example}
\end{table*}

\subsection{Results on All Languages}
\label{sec:compare}

\subsubsection{WER Comparison}
In Fig. \ref{fig:wer}, we show a large-scale comparison between the 2B baseline model and shallow fusion (SF) with three different LMs: Conf-140M, Conf-640M, and GLaM-64E models for 50 languages. We see that the all three LMs improve the baseline model for a number of languages, with regressions on a few languages (less than 10). The improvement increases as we increase the size of the dense LM from 140M to 640M, and further when we switch from a dense LM to a sparse model (GLaM-64E). Overall, the average WERs for the 2B baseline, Conf-140M SF, Conf-640M SF, and GLaM-64E SF are 8.50\%, 8.46\%, 8.38\%, and 8.22\%, respectively. Compared to the baseline, the Conf-140M, Conf-640M, and GLaM-64E improve the WER for 58\% (29/50), 76\% (38/50), and 86\% (43/50) languages, respectively. The per-language WER reduction is up to 16.3\% relative for GLaM-64E. On the other hand, we also note that shallow fusion leads to regression for a few languages. For example, there are WER regressions for GLaM-64E for fa-IR, hi-IN and mr-IN: 2.1\%, 0.6\%, and 2.2\%, respectively. However, the scale of regression is relatively small.

To better visualize the improvement of GLaM-64E compared to the baseline and other LMs, we show the relative WER reduction (WERR) of the GLaM-64E compared to the 2B baseline, Conf-140M, and Conf-640M, respectively, in Fig. \ref{fig:werr}. Compared to the baseline model, GLaM improves 86\% (43/50) languages, and the average WERR for the improved languages is 5.53\% relative. Compared to the 140M and 640M Conformer LMs, the GLaM-64E performs better for 82\% (41/50) and 68\% (34/50) languages, respectively. The average relative WERR for the improved languages are 3.85\% and 3.06\%, respectively. In terms of computation, the GLaM-64E only activates 2 experts in inference due to the sparsity design, and the computation is similar to the Conf-140M model. To our best knowledge, this is the first time we see a sparse MoE LM outperforms its state-of-the-art dense counterpart in large-scale multilingual shallow fusion.

\subsubsection{Decoding Examples}
To understand the types of improvements, we show some decoding examples between shallow fusion using GLaM-64E and Conformer-140M in Table \ref{tab:decoding_example}. We see that the GLaM wins mainly focus on person names and proper nouns. There is one interesting example that for en\_ph a code-switching example with a transliterated Chinese song name ``ni yao de ai" is recognized correctly in GLaM shallow fusion but not using Conformer. This indicates that a token-level MoE model may provide more flexibility in transliterating the language on the token level. Another win in cmn\_hant\_tw also indicates GLaM may have provided more diversity in modeling. On the other hand, GLaM suffers losses from over-correction (en\_au and cmn\_hant\_tw losses) and text normalization, even the over-correction in en\_au may seem to be reasonable. In the cmn\_hant\_tw example, code-switching data may be needed to further train the model to avoid code-switching errors. 

\section{Conclusion}
\label{sec:conclude}
We have explored incorporating large LMs in large-scale multilingual shallow fusion. For the first time, we have 1) scaled up the language coverage to 84 for shallow fusion, and 2) scaled up the LM from 140M dense Conformer LM to a 1.9B GLaM. We show that GLaM shallow fusion performs better than its dense counterparts, a 140M Conformer LM, and also outperforms a 640M Conformer LM. The improvement is achieved for at least 68\% of languages, with regression on less than 5 languages. Although we may need to store the 1.9GB GLaM on disk, the inference computation is similar to a 140M dense LM due to its MoE architecture.

\bibliographystyle{IEEEbib}
\renewcommand{\bibsection}{\section{REFERENCES}}
{\footnotesize\bibliography{refs}}

\end{document}